# A Unified Representation for Continuity and Discontinuity: Syntactic and Computational Motivations

**Ratna Kandala**[1,*]
n038k926@ku.edu
Department of Psychology,
University of Kansas, USA

**Prakash Mondal**[2]
prakashmondal@la.iith.ac.in
Department of Liberal Arts,
Indian Institute of Technology Hyderabad, India

**Abstract:** This paper advances a unified representation of linguistic structure for three grammar formalisms, namely, Phrase Structure Grammar (PSG), Dependency Grammar (DG) and Categorial Grammar (CG) from the perspective of syntactic and computational complexity considerations. The *correspondence principle* is proposed to enable a unified representation of the representational principles from PSG, DG, and CG. To that end, the paper first illustrates a series of steps in achieving a unified representation for a discontinuous subordinate clause from Turkish as an illustrative case. This affords a new way of approaching discontinuity in natural language from a theoretical point of view that unites and integrates the basic tenets of PSG, DG, and CG, with significant consequences for syntactic analysis. Then this paper demonstrates that a unified representation can simplify computational complexity with regards to the neurocognitive representation and processing of both continuous and discontinuous sentences vis-à-vis the basic principles of PSG, DG, and CG.

# 1 Introduction

Discontinuity refers to a case of non-adjacency when a predicate and its argument(s) are not adjacent as per the linear order of the sentence—predicate structure here may apply to constituents such as verb phrases, noun phrases, adjective phrases, etc. It is typically observed in free word order languages including Australian languages such as Warlpiri, Jiwarli, Turkish (Hale, 1982, 1983; Nordlinger, 2014). Figure 1 depicts a schematic representation of continuity and discontinuity.

---

[*] This work was done at IITH

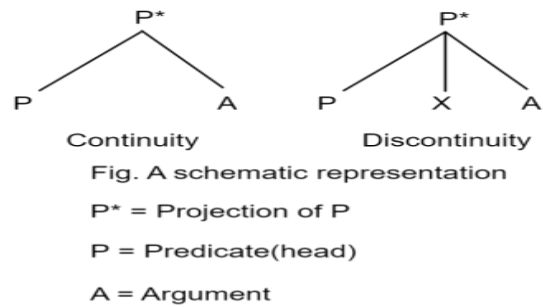

Fig. A schematic representation
P* = Projection of P
P = Predicate(head)
A = Argument

Fig.1. A schematic representation of discontinuity

There are divergent existing approaches towards discontinuity and most of them introduce ad hoc assumptions or constraints. For instance, Kaplan and Bresnan's (1982) work on Warlpiri has proposed a flat phrase structure by attributing it to the free base-generation of elements within the clause. Interestingly, McCawley (1982, 1987) has relaxed two important tree-organizational principles, namely, the *no-crossing constraint* (which bars crossing tree branches) and the *single mother condition* (which bars the sharing of a node by two mother nodes). This results in the 'tangling' of trees for discontinuous structures[†]. A similar kind of solution has also been proposed by Citko (2011) in an approach called *Parallel Merge* that relaxes the *single root condition* (comparable to the *single mother condition*). From a related perspective, Dowty (1996) has proposed a 'minimalist theory of syntax' which considers linear structure as the norm, instead of hierarchical structure. He also advances *unordered lists* – constituents that are more tightly bound to adjacent words—in order to account for discontinuity. In a similar vein, Donohue and Sag (1999) have accounted for discontinuity by proposing the 'sequence union operation' of two lists. For instance, the sequence union of two lists $l_1 = $ <p,q> and $l_2 = $ <r,s> is the list $l_3$ iff each of the elements in $l_1$ and $l_2$ is present in $l_3$ and the original order of the elements in $l_1$ and $l_2$ is preserved. As such, the sequence union of $l_1$ and $l_2$ is any of the following lists/sequences: <p,q,r,s>, <p,r,s,q>, <p,r,q,s>, <r,s,p,q>, <r,p,s,q>, <r,p,q,s> but not <q,p,r,s>, <p,q,s,r>, etc.

These current solutions, relying on certain ad hoc assumptions to tackle the issue of discontinuity, contrast with other previous attempts to integrate (the representational principles of) two grammar formalisms in specific ways. For instance, Bar-Hillel et al.

---

[†] It may be noted that if a line is drawn to connect X to its actual predicate outside the structure of P* (that is, a predicate other than P*) in Figure 1, the lines will be crossed.

(1960) have shown weak equivalence between PSG and CG. On the other hand, instead of relaxing PSG principles like McCawley, Barry and Pickering (1990, 1993, pp. 864-5) have argued for a flexible constituent called 'dependency constituent' in which all the words in a constituent are linked by dependencies. Along with the work of Barry and Pickering, other attempts to integrate constituency and dependency relations include Hays (1964, pp. 513), Robinson (1970, pp. 260-263) and Gaifman (1965, pp. 316-325). More recently, Dras et al. (2004) have drawn correspondences between dependency graphs and portions of auxiliary trees in Tree Adjoining Grammar (TAG). But auxiliary trees in TAG do not always correspond to PSG's constituents because auxiliary trees have exactly one frontier node marked as a foot node. In all, these solutions are limited towards showing correspondences between the representations of linguistic structure in two grammar formalisms.

The present paper presents a unique solution towards discontinuity in natural language. We propose a unified representation by integrating the (basic) representational principles of three grammar formalisms: Phrase Structure Grammar (PSG), Dependency Grammar (DG) and Categorial Grammar (CG), without stipulating *ad-hoc* constraints and unwarranted auxiliary assumptions. The unified representation will be rich enough to align with all the properties of the sentence that any of the three individual representations (in PSG, DG and CG) aligns with; in other words, the unified representation will do everything for us that any of the three individual representations does for us. By simply being anchored in the predicate-argument structure of constituents, the unified representation encodes the category of words, linear order, hierarchy/constituency relations, direct head-dependent relations, functor-argument relations among words in a sentence uniformly for both continuous and discontinuous structures—something that each of these grammar formalisms, taken in isolation, does not account for. We argue for its computational feasibility given the nature of language processing in our neurocognitive system.

This paper is organized as follows: Section 2 provides a brief background to the three grammar formalisms; section 3 discusses the significance of *the correspondence principle* to be used in unifying DG and CG representations. Section 4 illustrates the derivations of a discontinuous subordinate clause from Turkish as an illustrative case. Section 5 offers the consequences for the analysis of certain syntactic phenomena. Section 6 provides a discussion on the cognitive-computational grounding of the unified representation. Finally, the paper ends with concluding remarks on the implications of this work for further research in section 7.

## 2	Brief Overviews of PSG, DG and CG

In this section, we present a short introduction to each of the formalisms.

**2.1 Phrase Structure Grammar (PSG)**

The rise of transformational-generative grammar can be attributed to Chomsky's *Syntactic Structures* (1957) which introduced a structured way of analysis of sentences in which the notion of constituency is crucial to this formalism. In essence, a phrase structure grammar is characterized by a finite set of initial strings $\Sigma$: $\Sigma_1 \ldots \Sigma_m$ and a finite set of rules F of the form F: $X_1 \rightarrow Y_1 \ldots X_m \rightarrow Y_m$, where any X and Y are strings and Y is formed from X via the replacement of X with Y (Chomsky, 1956, 1965, pp. 111-112). Since such replacements via rewriting are supposed to be implemented in the absence of the structural contexts of X (that is, in possibly null contexts), the formalism of grammar is usually known as *context-free grammar*. In essence, PSG can be defined by a 4-tuple <N, T, P, S>, where N is the set of nonterminal symbols such as Det, NP, VP, etc., T is the set of terminal symbols, that is, words, P is the set of production/rewriting rules and S is the start symbol, a member of N, usually recognized as the symbol for a sentence/clause. As such, PSG can be viewed as a formal system with rules of the form A → B C D (Gazdar, 1983; Gazdar et. al., 1985). This rule specifies two relations: Dominance and Precedence. It states that A dominates B, C and D, and that B is the leftmost node while D is the rightmost node with C in between. For example, consider *the bright sun*. The rule NP → Det AP N states that NP dominates the three nodes, namely, Det, AP, N, and also that Det is the leftmost node and N is the rightmost node. Figure 2 depicts the PSG tree of the NP *the bright sun*.

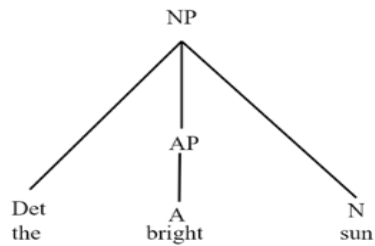

Fig 2. The PSG tree of the phrase *the bright sun*

Though there have been further developments of PSG such as X-Bar Theory, Minimalist Program (Chomsky, 1970, 1995), these core ideas of constituency, dominance, precedence and succession remain the same (Matthews, 1981; Leffel & Bouchard, 1991; Moro, 2008; Carnie, 2010; Everaert et al., 2015) and are taken into account for the present work.

**2.2 Dependency Grammar (DG)**

The formalism of DG developed by Tesnière (1959) analyses sentences based on head-dependent relations. It can be formally defined as a 4-tuple: DG = $<V_N, V_T, D, R>$, where $V_N$ is the set of auxiliary elements (syntactic categories), $V_T$ is the set of terminal elements (minimal syntactic units), D is the set of dependency rules and R is the initial symbol at the root of the tree (that is, $R \in V_N$). A dependency rule in D is a statement consisting of one auxiliary element functioning as the governing element or head (I) and any finite number of auxiliary elements as the dependents. There are two important rules in D: Rule 1: $I(D_1,...,D_m * D_i,...,D_k)$ (i, m, k $\geq 0$ ; not always i=m=k). Rule 2: I (*). Here, $D_1,...,D_m$ refers to the dependents to the left of I and $D_i,...,D_k$ are the dependents to the right of I. For example, the rule V(Det1,N1*Det2,N2) for *The girl likes the flower* says that V is the independent word, Det1 and N1 are the dependents to the left of V, and Det2 and N2 are the dependents to the right of V (Hays, 1964; Gaifman, 1965; Debusmann, 2000; de Marneffe & Nivre, 2019; Osborne, 2019). Figure 3 depicts a DG graph of the same clause. It is to be noted that DG as a theoretical framework in linguistics focuses on the concept of 'valency', which describes the number of arguments that a predicate requires. While DG theorists such as Hudson (1984, pp. 92) and Miller (2000, pp. 22) have adopted a subgraph or subtree-based notion of constituents, the fundamental idea of valency and dependency valuation remains central to DG. The dependency valuation function, denoted as δ, is a key aspect of DG, as it takes a node in a sentence as input and yields a real-valued output (Levelt 2008, pp. III 51). By comparing the values generated by the δ function for different nodes in a sentence, we can determine the hierarchical relationships that exist between them. For example, if node A is dependent on node B, that is, A~B, then δ(A) > δ(B). This function typically starts from 0 at the top of the tree but can also be adjusted to start from 1. The δ function will prove to be a useful tool for recoding the functor-argument relations in terms of dependency relations (in section 4). This will help arrive at a unified representation to be shown in the present paper.

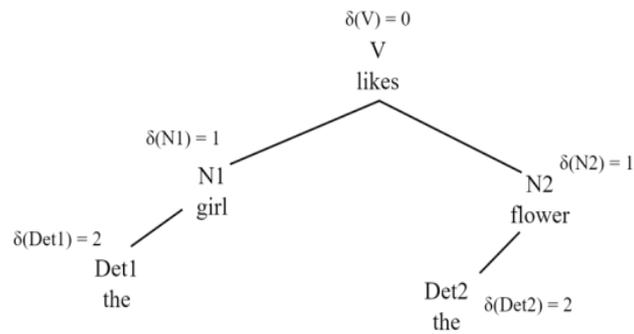

Fig 3. A DG graph of *The girl likes the flower*

## 2.3 Categorial Grammar (CG)

CG is a context-free grammar formalism developed by Bar-Hillel (1953). The term 'categorial' is derived from the notion of 'category' because words are assigned categories in terms of N and S in this formalism. It can be formally defined as a 4-tuple: <V, C, R, F> where V is the set of all lexical items in a language, C = {N, S}, R is the set of functional compositional rules for the generation of categories for lexical items, and F is the function that maps each lexical item to its set of categories. For example, in English, the category of a determiner *the* (note that *the* ∈ V) in the phrase *the girl* is N/N, meaning that it is a function that takes a noun as its input and outputs another noun, that is, *f*(N)= N. This is specified by R. On the other hand, if we have that F(*open*)= {N\S, (N\S)/N}, the first category in this set is for the intransitive use of *open*, while the second is for the transitive use. CG, thus, analyses sentences in terms of functor-argument relations (Steedman, 1992, 2014). We adopt Lambek's (1958) slash notation for the present work. Here, the X/Y category represents a function X which accepts an argument Y to its right. For example, N/N is a category representing a function N and accepts an argument of the category N to its right. Figure 4 depicts the CG derivation of *The girl likes the flower*.

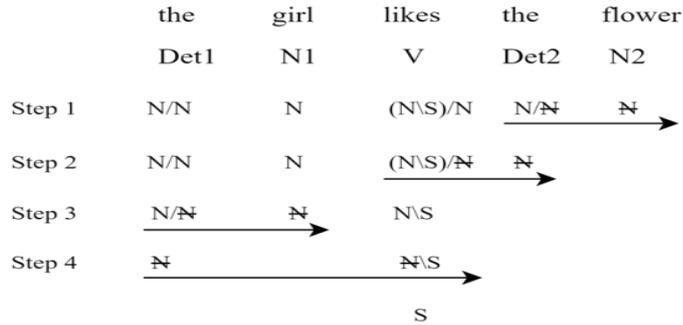

Fig 4. The CG derivation of *The girl likes the flower*

As we shall see, these three formalisms are not distinct as they are considered to be and the following principle can be applied to unify the basic representational principles of these three formalisms.

## 3      The Correspondence Principle and its Significance

For any given two words A and B, the head-dependent relations and the functor-argument relations can be unified by the following formulation, which we call *the correspondence principle* (Anonymous, 2022, 2023, 2024).

$$A(B^*) \vee A(^*B) \equiv A|B \qquad (1)$$

Here, the Left-Hand Side (LHS) specifies the DG relation(s) between A and B, and the Right-Hand Side (RHS) corresponds to the CG relation(s) between A and B. Specifically, $A(B^*)$ indicates that B is dependent on A and B is to the left of A; $A(^*B)$ indicates that B is dependent on A and B is to the right of A as per the linear order (from the viewer's perspective of the right and the left). '∨' is logical disjunction and '≡' is a special equivalence sign indicating the equivalence relation between DG representations on the left and CG representations on the right. The '|' sign on the RHS indicates the

neutral direction for a functor-argument relation between the two words A and B—either A or B can be the functor, or conversely, either A or B can be the argument. This formulation is a simplification and generalization of the equivalence relation between head-dependent and functor-argument relations between any two words A and B. To demonstrate the equivalence, we show that one can derive the functor-argument relation from the head-dependent relation between A and B and also vice versa in the DG → CG and CG → DG derivations. Given that categorial functor-argument relations can be easily defined on standard constituency relations, *the correspondence principle* ultimately helps draw correspondences between head-dependent relations, functor-argument relations and, derivatively, constituency relations (to be demonstrated in section 4).

Crucially, this principle implies that when one of A and B is the functor or the head, the role of the other element will be of the opposite category (as the argument or the dependent category) across sides. The logical relation is that of an *implication*, but not of *an entailment*. If A happens to be the head on the LHS, it does not necessarily have to play the role of the functor on the RHS, because the formulation does not say anything about B across sides when the role of A on either side is fixed (either as the head or as the functor), and also vice versa. Thus, if A happens to be the head word on the LHS and the functor on the RHS, B will be whatever the dependent category of A is on the LHS and whatever the argument of A is on the RHS. Conversely, if A happens to be the dependent word on the LHS and the argument on the RHS, B will be whatever the head of A is on the LHS and whatever the functor of A is on the RHS. More concretely, in cases where there is a direct dependency relation between the functor and the argument, A and B on the Left-Hand Side (LHS) and Right-Hand Side (RHS) turn out to be the same. However, this is not the case always. In exceptional cases, only one of A and B tends to be the same on the LHS and RHS, and the other category can vary across sides. If, for example, we suppose that A is the same on both sides, the exact value or role of B may differ on the LHS and the RHS (that is, B can take a word X, for example, on the LHS, while it takes a word Y, for example, on the RHS).

Notably, previous solutions towards discontinuity such as McCawley's 'tangling' of trees (1982) and Citko's (2011) 'parallel merge' approach relax PSG's *no-crossing* *constraint* and the *single mother condition*. However, this appears to be an *ad-hoc* solution and also creates multiple rules for sentences in continuous and discontinuous forms. In contrast, in the context of the present work, the relaxation of the *no-crossing constraint* and the *single mother condition* is the very expression or manifestation of *the correspondence principle*, which forms the basis of the relaxation of the *no-crossing constraint* and the *single mother condition*. Tangling of trees and sharing of a node by two mother nodes due to discontinuity can be viewed as macro-level side effects of the unification of functor-argument and head-dependent relations at local levels of organization (that is, at the level of words and phrases). This will become clearer in section 4 below.

## 4 The Unified Representation: A Discontinuous Subordinate Clause as an Illustration

In this section, we discuss the way in which a unified representation can be achieved for a discontinuous subordinate clause from Turkish. We begin by depicting the CG derivation in the PSG tree (PSG → CG derivation). In the next step, for each functor-argument relation, we draw the corresponding PSG tree depicting the constituent structure (CG → PSG derivation). Then, we show how one can reformulate the direct head-dependent relations in terms of the functor-argument relations in harmony with the CG derivation (DG → CG derivation). Afterwards, we establish the equivalence relation between CG and DG by deriving the dependency relations from the CG relations (CG → DG derivation). The following is a Turkish subordinate clause:

1. *hemen   gel-iyor-um      diye      git-ti          dön-me-di*.
    soon come-PROG-1SG  diye      go-PAST.3SG  come.back-NEG-PST.3SG
   '(S)He went saying "I will come soon" but (s)he didn't come back.'

It is to be noted that, here, *diye* functions like a subordinator (Gündoğdu, 2017, pp. 35). Firstly, let us consider the PSG → CG derivation.

**(4a) The CG derivation in the phrase structure tree (PSG → CG)**

Figure 5 depicts the CG derivation of (1) in its PSG tree whose tangling needs to be understood in terms of the unification of PSG, DG, and CG relations to be shown in (4a-d) below and Figure 6 depicts the CG derivation of (1).

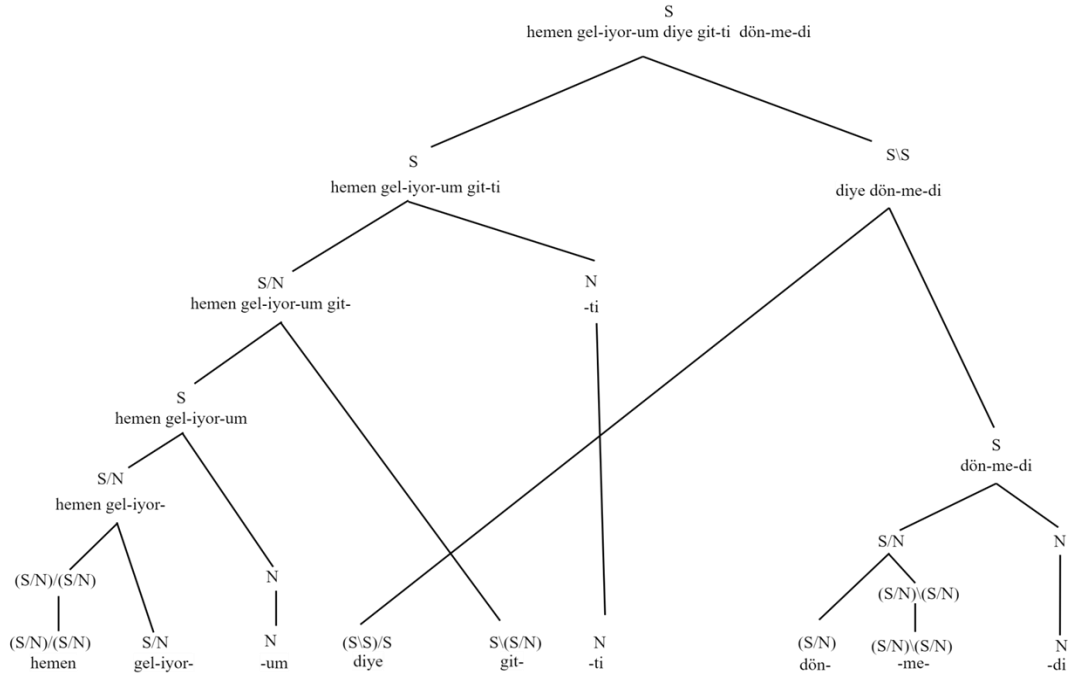

Fig 5. The CG derivation in a PSG tree of (1)

|  | hemen | gel-iyor- | -um | diye | git- | -ti | dön- | -me- | -di |
|---|---|---|---|---|---|---|---|---|---|
|  | Adv | V1 | N1 | Conj | V2 | N2 | V3 | Neg | N4 |
| Step 1 | (S/N)/(S/N) | (S/N) | N | (S/S)/S | S\(S/N) | N | ~~(S/N)~~ | ~~(S/N)~~\(S/N) | N |
| Step 2 | (S/N)/(S/N) | (S/N) | N | (S/S)/S | S\(S/N) | N |  | S/~~N~~ | ~~N~~ |
| Step 3 | (S/N)/(S/N) | (S/N) | N | (S/S)/~~S~~ | S\(S/N) | N |  | ~~S~~ |  |
| Step 4 | (S/N)/~~(S/N)~~ | ~~(S/N)~~ | N | S/S | S\(S/N) | N |  |  |  |
| Step 5 | S/~~N~~ |  | ~~N~~ | S/S | S\(S/N) | N |  |  |  |
| Step 6 | ~~S~~ |  |  | S/S | ~~S~~\(S/N) | N |  |  |  |
| Step 7 |  |  |  | S/S | S/~~N~~ | ~~N~~ |  |  |  |
| Step 8 |  |  |  | S/~~S~~ | ~~S~~ |  |  |  |  |
|  |  |  |  | S |  |  |  |  |  |

Fig 6. The CG derivation of (1)

*The illustration of Fig 6:*

- In step 1, the category of *dön-* is cancelled out with respect to the category of *-me-*. This step builds the meaning conveyed through *didn't come back*.
- In step 2, the category of *-di* is cancelled out with respect to the category of *-me*. This step builds the meaning conveyed through *(s)he didn't come back*.
- In step 3, the category of *-me-* is cancelled out with respect to the category of *diye*. This step builds the meaning conveyed through *but (s)he didn't come back*.
- In step 4, the category of *gel-iyor-* is cancelled out with respect to the category of *hemen*. This step builds the meaning conveyed through *will come soon*.
- In step 5, the category of *-um* is cancelled out with respect to the category of *hemen*. This step builds the meaning conveyed through *I will come soon*.
- In step 6, the category of *hemen* is cancelled out with respect to the category of *git-*. This step builds the meaning conveyed through *went saying "I will come soon"*.
- In step 7, the category of *-ti* is cancelled out with respect to the category of *git-*. This step builds the meaning conveyed through *(s)he went saying "I will come soon"*.
- In step 8, the category of *git-* is cancelled out with respect to the category of *diye*. This step builds the meaning conveyed through *(S)He went saying "I will come soon" but (s)he didn't come back*.

It needs to be emphasized here that the CG derivations in discontinuous sentences proceed normally when they involve functors whose argument(s) are non-adjacent. The only proviso here is that the interpretation of the functor-argument relation in the case of discontinuity can be permitted by the wrapping[‡] operation (Morrill, 1995, pp. 197-98; Steedman, 2014, pp. 682-683). This may allow a discontinuous element to be taken as an argument by a functor separated by some intervening expression. For instance, the category of V2 as a functor may take that of Adv as an argument over the intervening category of Conj. under wrapping in step 6 of Fig 6, although this is not explicitly shown there. Similar considerations may apply to step 3 of Fig 6, where the category of Conj. can take that of Neg over the combined form of V2 and N2. The special property of wrapping permits the construction of a discontinuous product such that two elements can combine over another intervening element so that the combined form can undergo a functor-argument relation with the intervening element (Morrill et al., 2011). In fact, nothing prevents wrapping as an operation from being applied iteratively in order that it can cover cases where two elements are separated by more than one intervening element. This is what we shall assume for CG derivations in this paper. Also, we have not adopted type-raising operation to avoid arbitrary complexity in the CG derivations and hence we have adopted Lambek's basic notations in functor-argument formulations. Now, let us consider the converse derivation of the previous one, that is, the CG → PSG derivation.

---

[‡] Wrapping rules usually infix, by way of a sort of swapping, a discontinuous string element in a place where another element was initially located (see for details, Steedman, 1985).

**(4b) The CG → PSG derivation**

Below are the steps in the formation of the final PSG tree (Figure 13) from the corresponding steps of the CG derivation (Figure 6).

Step 1 of the CG derivation builds the meaning *didn't come back* conveyed by the VP *dön-me-* as seen in Figure 7.

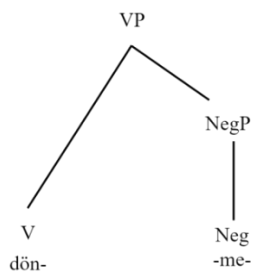

Fig 7. The PSG tree corresponding to step 1 of the CG derivation

Step 2 of the CG derivation builds the meaning *(s)he didn't come back* conveyed by the S *dön-me-di*, as seen in Figure 8.

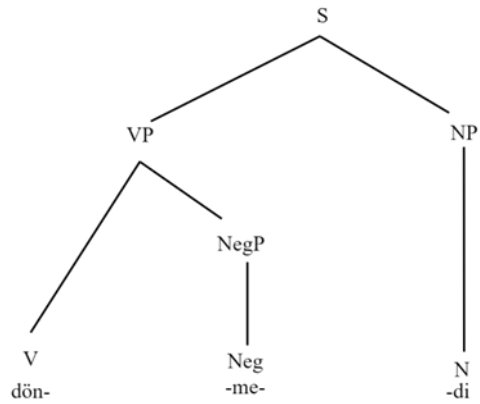

Fig 8. The PSG tree corresponding to step 2 of the CG derivation

Step 3 of the CG derivation builds the meaning *but (s)he didn't come back* conveyed by the S *diye dön-me-di*, as seen in Figure 9.

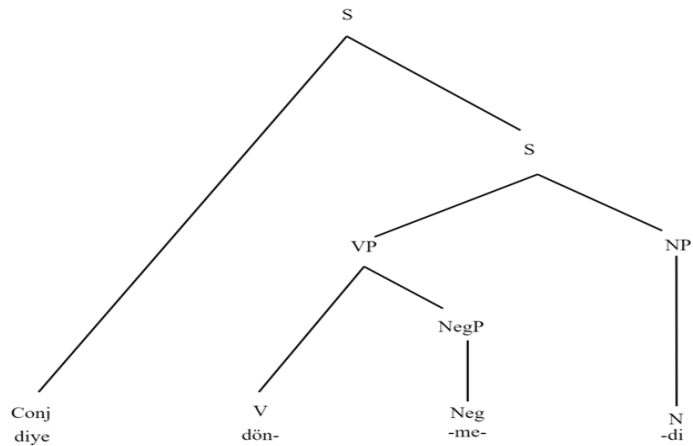

Fig 9. The PSG tree corresponding to step 3 of the CG derivation

Step 4 of the CG derivation builds the meaning *will come soon* conveyed by the VP *hemen gel-iyor-*, as seen in Figure 10.

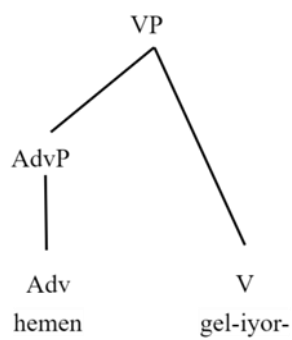

Fig 10. The PSG tree corresponding to step 4 of the CG derivation

Step 5 of the CG derivation builds the meaning *I will come soon* conveyed by the S *hemen gel-iyor-um*, as seen in Figure 11.

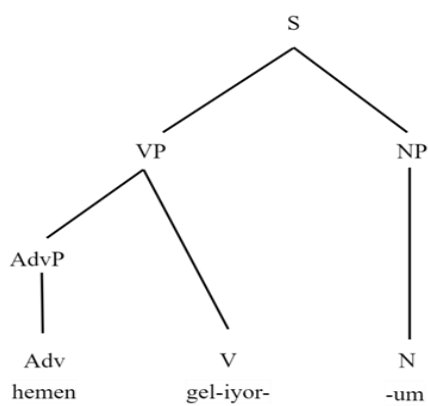

Fig 11. The PSG tree corresponding to step 5 of the CG derivation

Step 6 of the CG derivation builds the meaning *went saying "I will come soon"* conveyed by the VP *hemen gel-iyor-um git-*, as seen in Figure 12.

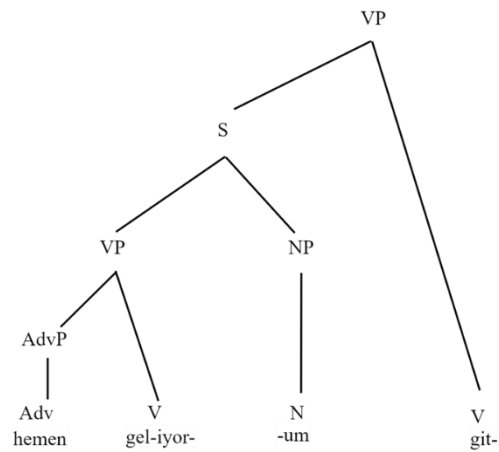

Fig 12. The PSG tree corresponding to step 6 of the CG derivation

Step 7 of the CG derivation builds the meaning *(s)he went saying "I will come soon"* conveyed by the S *hemen gel-iyor-um git-ti*, as seen in Figure 13.

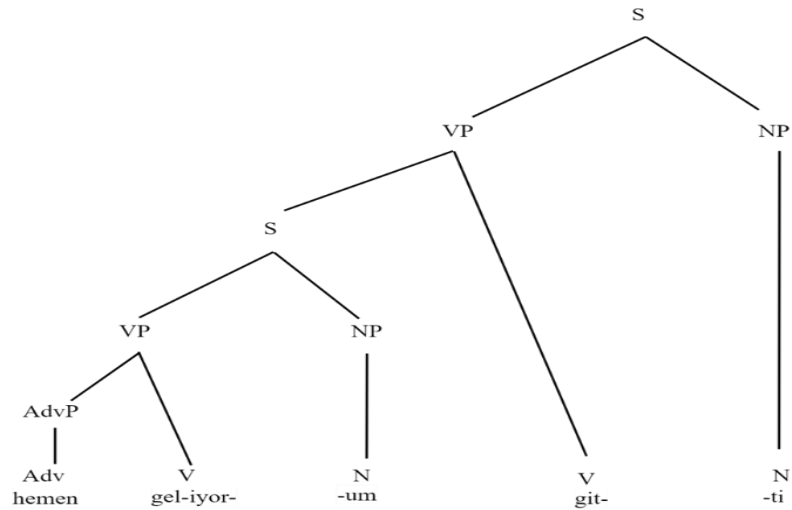

Fig 13. The PSG tree corresponding to step 7 of the CG derivation

Step 8 of the CG derivation builds the meaning *(S)He went saying "I will come soon" but (s)he didn't come back* conveyed by the S *hemen gel-iyor-um diye git-ti dön-me-di*, as seen in Figure 14.

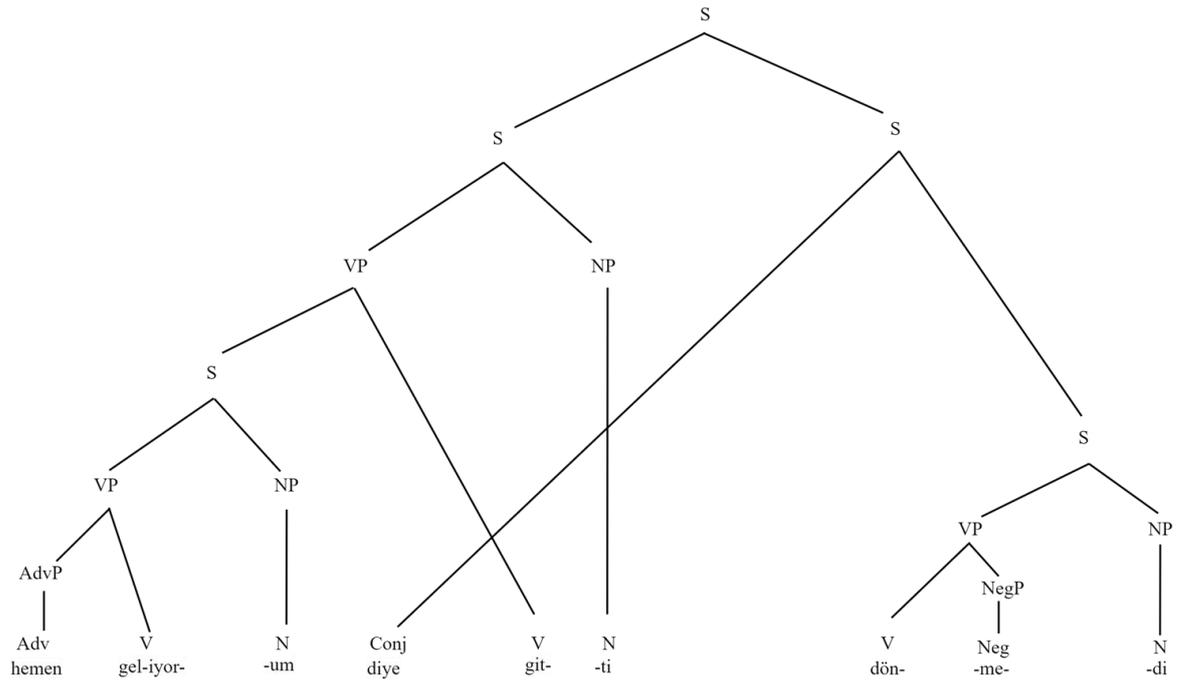

Fig 14. The final PSG tree

Accordingly, the corresponding PSG rules are: S → S S ; S → VP NP ; S → Conj S; VP → S V; VP → V NegP ; VP → AdvP V; NP → N; AdvP → Adv; NegP → Neg.

The two derivations above have established an equivalence relation between PSG and CG from both the directions (PSG→CG and CG→PSG). Next, let us move on to the third step, that is, the DG → CG derivation.

**(4c) Dependency functions in terms of CG formulae (DG → CG)**

Here, for each head-dependent relation the corresponding functor-argument relation is written using the dependency function (that is, the δ function). When there is no direct head-dependent relation between the functor and its argument, *the correspondence principle* is used to show how one can rewrite the direct dependency relation between the head and its dependent in terms of the functor-argument relation for that particular step of the CG derivation. The dependency graph for sentence (1) is shown below.

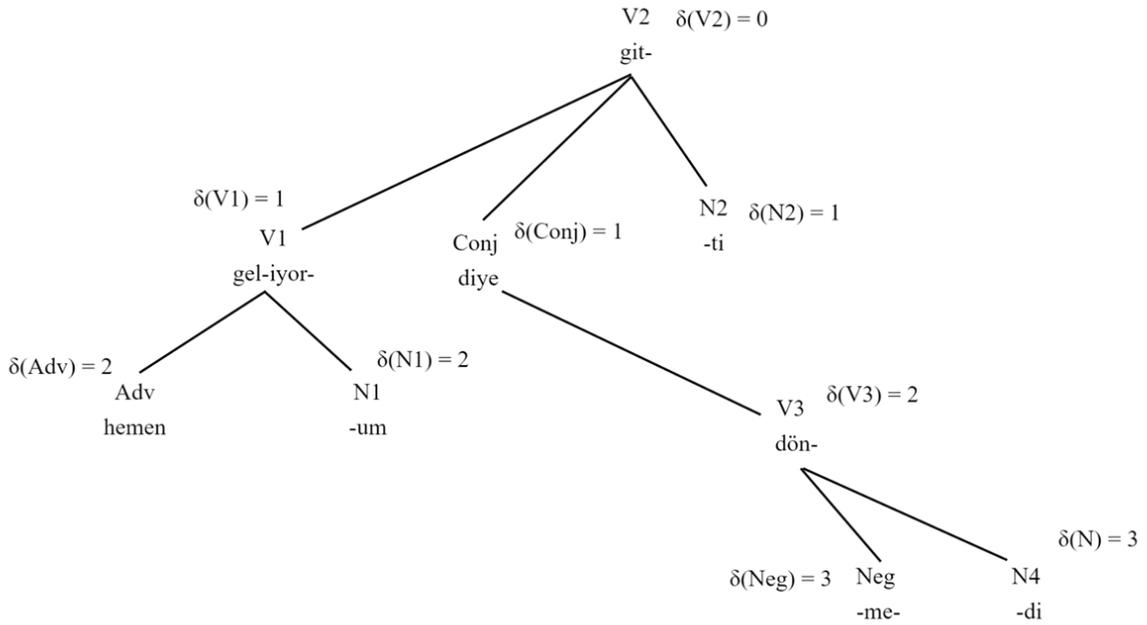

Fig 15. The DG graph of (1)

Figure 15 illustrates the following dependencies:

1. V3(*Neg)   or *-me-* is dependent on *dön-*
2. V3(*N4)    or *-di* is dependent on *dön-*
3. Conj(*V3)  or *dön-* is dependent on *diye*
4. V1(Adv*)   or *hemen* is dependent on *gel-iyor-*
5. V1(*N1)    or *-um* is dependent on *gel-iyor-*
6. V2(V1*)    or *gel-iyor-* is dependent on *git-*
7. V2(*N2)    or *-ti* is dependent on *git-*
8. V2(Conj*)  or *diye* is dependent on *git-*

Here *the correspondence principle* would help unite the DG and CG representations. The following steps correspond to those in the CG derivation in Figure 6:

Step 1: ~~δ(V3)~~   δ(V3)\δ(Neg)

This step captures the functor-argument relation between *-me-* (Neg) and *dön-* (V3) corresponding to step 1 of the CG derivation in Figure 6. This step builds the meaning conveyed through *didn't come back*.

Step 2:  δ(V3)/~~δ(N4)~~ ~~δ(N4)~~

This step captures the functor-argument relation between *-me-* (Neg) and *-di* (N4) corresponding to step 2 of the CG derivation in Figure 6. By using *the correspondence principle* we have *dön-*(*-di*) ≡ *-me-*/*-di*. In other words, V3(*N4) ≡ Neg/N4. This can also be expressed as B(*A) ≡ B/A. Here, A corresponds to *-di* (N4) on both the RHS and the LHS. B on the LHS corresponds to *dön-* (V3) and B on the RHS corresponds to *-me-* (Neg). Since *-me-* (Neg) and *-di* (N4) do not participate in any (direct) dependency relation as seen in Figure 3, the functor-argument relation is constructed through *dön-* (V3) and *-di* (N4). This step builds the meaning conveyed through *(s)he didn't come back*.

Step 3:  δ(Conj)/~~δ(V3)~~ ~~δ(V3)~~

This step captures the functor-argument relation between *diye* (Conj) and *-me-* (Neg) corresponding to step 3 of the CG derivation in Figure 6. By using *the correspondence principle*, the functor-argument relation can be constructed through *diye* (Conj) and *dön-* (V3) because no direct dependency relation exists between *diye* (Conj) and *-me-* (Neg). Therefore, *diye*(**dön-*) ≡ *diye*/*-me-*. This can also be written as Conj(*V3) ≡ Conj/Neg or A(*B) ≡ A/B. Here A corresponds to *diye* (Conj) on both the LHS and RHS. B on the LHS corresponds to *dön-*(V3) and B on the RHS corresponds to *-me-* (Neg). This step builds the meaning conveyed through *but (s)he didn't come back*.

Step 4: δ(Adv)/~~δ(V1)~~ ~~δ(V1)~~

This step captures the functor-argument relation between *hemen* (Adv) and *gel-iyor-* (V1) corresponding to step 4 of the CG derivation in Figure 6. This step builds the meaning conveyed through *will come soon*.

Step 5:  δ(V1)/~~δ(N1)~~ ~~δ(N1)~~

This step captures the functor-argument relation between *hemen* (Adv) and *-um* (N1) corresponding to step 5 of the CG derivation. Here, a direct dependency relation does not exist between the argument *-um* (N1) and the functor *hemen* (Adv). Therefore, the functor-argument relation can be constructed through the direct dependency relation between *gel-iyor-* (V1) and *-um* (N1) using *the correspondence principle*: *gel-iyor-*(*- *um*) ≡ *hemen*/-*um*. This can also be expressed as V1(*N1) ≡ Adv/N1 or B(*A) ≡ B/A. Here, A corresponds to -*um* (N1) on both the RHS and the LHS. B on the RHS corresponds to *hemen* (Adv) and B on the LHS corresponds to *gel-iyor-* (V1). This step builds the meaning conveyed through *I will come soon*.

Step 6: ~~δ(V1)~~ δ(V1)\δ(V2)

This step captures the functor-argument relation between *hemen* (Adv) and *git-* (V2) corresponding to step 6 of the CG derivation in Figure 6. A direct dependency relation does not exist between the argument *hemen* (Adv) and the functor *git-* (V2); instead, *gel-iyor-* (V1) is directly dependent on *git-* (V2). Hence using *the correspondence principle,* we have *git-*(*gel-iyor-**) ≡ *hemen*\\*git-* or V2(V1*) ≡ Adv\V2 or A(B*) ≡ B\A. Here, A corresponds to *git-* (V2) on both the RHS and the LHS. B on the RHS corresponds to *hemen* (Adv) and B on the LHS corresponds to *gel-iyor-* (V1). This step builds the meaning conveyed through *went saying "I will come soon"*.

Step 7: δ(V2)/~~δ(N2)~~ ~~δ(N2)~~

This step captures the functor-argument relation between *git-* (V2) and *-ti* (N2) corresponding to step 7 of the CG derivation in Figure 6. This step builds the meaning conveyed through *(s)he went saying "I will come soon"*.

Step 8: δ(Conj)/~~δ(V2)~~ ~~δ(V2)~~

This step captures the functor-argument relation between *diye* (Conj) and *git-* (V2) corresponding to step 8 of the CG derivation in Figure 6. This step builds the meaning conveyed through the entire S: *(S)He went saying "I will come soon" but (s)he didn't come back.*

Let us now consider the converse derivation of the previous one, that is, the CG → DG derivation.

**(4d) The CG → DG derivation**

Here *the correspondence principle* is used to show how the dependency relations can be derived from the categories assigned to the words and the subsequent CG derivation. When each step of the CG derivation is taken into account and expressed in terms of head-dependent relations, the corresponding dependency relation between the functor and the argument can be established.

*Step 1: The CG relation between 'dön-' and '-me-'*

In this relation, *-me-* (Neg) is the functor and *dön-* (V3) is the argument. The direction of the argument is to the left. Accordingly, the RHS of *the correspondence principle* would be *dön-\-me-* or V3\Neg. If we consider V3 to be A and Neg to be B, the RHS can also be written as A\B. There is a direct dependency relation between the functor and the argument, with *dön-* (V3) as the head and *-me-* (Neg) as its dependent. Hence the LHS of *the correspondence principle* would be: *dön-(\*-me-)* or V3(\*Neg) or A(\*B). Thus, the equivalence relation for step 1 of the CG derivation would be: *dön-(\*-me-)* ≡ *dön-\-me-*. This can also be expressed as: V3(\*Neg) ≡ V3\Neg or A(\*B) ≡ A\B.

*Step 2: The CG relation between '-me-' and '-di'*

In this CG relation, *-me-* (Neg) is the functor and *-di* (N4) is the argument. The direction of the argument is to the right. If we consider N4 to be A and Neg to be B, the RHS of *the correspondence principle* indicating the functor-argument relation between the two words would be: *-me-/-di* or Neg/N4 or B/A. If we turn to the LHS, there is no direct head-dependent relation between the functor and the argument; rather, *-di* (N4) is dependent on *dön-* (V3). Since the functor and the argument do not participate in any direct head-dependent relation, considering *-di* (N4) to be A on the RHS implies that B on the LHS is *dön-* (V3) and B on the RHS is *-me-* (Neg). Thus, the equivalence relation for step 2 of the CG derivation would be: *dön-(\*-di)* ≡ *-me-/-di*. This can also be expressed as: V3(\*N4) ≡ Neg/N4 or B(\*A) ≡ B/A. This relation clearly shows that *–di* (N4) is directly dependent on *dön-* (V3) but is the functor of *-me-* (Neg).

*Step 3: The CG relation between 'diye' and '-me-'*

In this CG relation, *diye* (Conj) is the functor and *-me-* (Neg) is the argument. The direction of the argument is to the right. If we consider *diye* (Conj) to be A and *-me-* (Neg) to be B, the RHS of *the correspondence principle* indicating the functor-argument relation between the two words would be: *diye/-me-* or Conj/Neg or A/B. If we speak of the LHS, there is no direct head-dependent relation between the functor and the argument; rather, *dön-* (V3) is dependent on *diye* (Conj). Since the functor and the argument do not participate in any direct head-dependent relation, considering *diye* (Conj) to be A on RHS would imply that B on the LHS is *dön-* (V3) and B on the RHS is *-me-* (Neg). Thus the equivalence relation for step 3 of the CG derivation would be: *diye*(\**dön-*) ≡ *diye*/*-me-*. This can also be expressed as: Conj(*V3) ≡ Conj/Neg or A(*B) ≡ A/B. This relation clearly shows that *diye* (Conj) is the head of *dön-* (V3) but functor of *-me-* (Neg).

*Step 4: The CG relation between 'hemen' and 'gel-iyor-'*
In this CG relation, *hemen* (Adv) is the functor and *gel-iyor-* (V1) is the argument. The direction of the argument is to the right. Accordingly, the RHS of *the correspondence principle* would be *hemen*/*gel-iyor-* or Adv/V1. If we consider Adv to be A and V1 to be B, the RHS can also be written as A/B. There is a direct dependency relation between the functor and the argument, with *gel-iyor-* (V1) as the head and *hemen* (Adv) as its dependent. Hence the LHS of *the correspondence principle* would be: *gel-iyor-*(*hemen**) or V1(Adv*) or B(A*). Thus, the equivalence relation for step 4 of the CG derivation would be: *gel-iyor-*(*hemen**) ≡ *hemen*/*gel-iyor-*. This can also be expressed as: V1(Adv*) ≡ Adv/V1 or B(A*) ≡ A/B.

*Step 5: The CG relation between 'hemen' and '-um'*
In this CG relation, *hemen* (Adv) is the functor and *-um* (N1) is the argument. The direction of the argument is to the right. If we consider *-um* (N1) to be A and *hemen* (Adv) to be B, the RHS of *the correspondence principle* indicating the functor-argument relation between them would be: *hemen*/*-um* or Adv/N1 or B/A. However, there is no direct head-dependent relation between the functor and the argument; rather, *-um* (N1) is dependent on *gel-iyor-* (V1). Accordingly, considering *-um* (N1) to be A would indicate B on the RHS is its functor *hemen* (Adv) and B on the LHS is its head *gel-iyor-* (V1). Hence the equivalence relation for step 5 of the CG derivation would be: *gel-iyor-*(\**-um*) ≡ *hemen*/*-um*. This can also be expressed as: V1(*N1) ≡ Adv/N1 or B(*A) ≡ B/A.

*Step 6: The CG relation between 'hemen' and 'git-'*

In this CG relation, *git-* (V2) is the functor and *hemen* (Adv) is the argument. The direction of the argument is to the left. If we consider *git-* (V2) to be A and *hemen* (Adv) to be B, the RHS of *the correspondence principle* indicating the functor-argument relation between them would be: *hemen\git-* or Adv\V2 or B\A. However, there is no direct head-dependent relation between the functor and the argument; rather, *gel-iyor-* (V1) is directly dependent on *git-* (V2). Accordingly, considering *git-* (V2) to be A would indicate that B on the RHS is *hemen* (Adv) and B on the LHS is *gel-iyor-* (V1). Hence the equivalence relation for step 6 of the CG derivation would be: *git-*(*gel-iyor-\**) ≡ *hemen\git-*. This can also be expressed as: V2(V1*) ≡ Adv\V2 or A(B*) ≡ B\A. This relation clearly shows that *git-* (V2) is the head of *gel-iyor-* (V1) but the functor of *hemen* (Adv).

*Step 7: The CG relation between 'git-' and '-ti'*

In this CG relation, *git-* (V2) is the functor and *-ti* (N2) is the argument. The direction of the argument is to the right. If we consider *git-* (V2) to be A and *-ti* (N2) to be B, the RHS of *the correspondence principle* indicating the functor-argument relation between them would be: *git-/-ti* or V2/N2 or A/B. Since there is a direct head-dependent relation between the functor and the argument, with *git-* (V2) as the head and *-ti* (N2) as its dependent, the LHS of *the correspondence principle* would be *git-*(*\*-ti*) or V2(*N2) or A(*B). Therefore, the equivalence relation for step 7 of the CG derivation would be: *git-*(*\*-ti*) ≡ *git-/-ti*. This can also be expressed as: V2(*N2) ≡ V2/N2 or A(*B) ≡ A/B.

*Step 8: The CG relation between 'diye' and 'git-'*

In this CG relation, *diye* (Conj) is the functor and *git-* (V2) is the argument. The direction of the argument is to the right. Accordingly, the RHS of *the correspondence principle* would be *diye/git-* or Conj/V2. If we consider Conj to be A and V2 to be B, the RHS can also be written as A/B. There is a direct dependency relation between the functor and the argument, with *git-* as the head and *diye* as the dependent. Hence the LHS of *the correspondence principle* would be: *git-*(*diye\**) or V2(Conj*) or B(A*). Thereby, the equivalence relation for step 8 of the CG derivation would be: *git-*(*diye\**) ≡ *diye/git-*. This can also be expressed as: V2(Conj*) ≡ Conj/V2 or B(A*) ≡ A/B.

Thus, combining the DG relations of all the steps would give us the DG graph (Figure 15). As we have established the desired equivalence between (i) PSG and CG and, then, between (ii) DG and CG in both forward and converse directions, the final step is to show that the unified representation can be achieved by combining all the above four derivations.

**(4e) A unified representation (from DG → CG → PSG)**

The unified representation for the Turkish sentence (1) is depicted in Figure 16.

| | hemen | gel-iyor- | -um | diye | git- | -ti | dön- | -me- | -di |
|---|---|---|---|---|---|---|---|---|---|
| | Adv | V1 | N1 | Conj | V2 | N2 | V3 | Neg | N4 |
| Step 1 | δ(Adv)/δ(V1) | δ(V1) | δ(N1) | [δ(Conj)/δ(V2)]/δ(V3) | δ(V1)\[δ(V2)/δ(N2)] | δ(N2) | ~~δ(V3)~~ | ~~δ(V3)~~\δ(Neg) | δ(N4) |
| Step 2 | δ(Adv)/δ(V1) | δ(V1) | δ(N1) | [δ(Conj)/δ(V2)]/δ(V3) | δ(V1)\[δ(V2)/δ(N2)] | δ(N2) | δ(V3)/~~δ(N4)~~ | | ~~δ(N4)~~ |
| Step 3 | δ(Adv)/δ(V1) | δ(V1) | δ(N1) | [δ(Conj)/δ(V2)]/~~δ(V3)~~ | δ(V1)\[δ(V2)/δ(N2)] | δ(N2) | ~~δ(V3)~~ | | |
| Step 4 | δ(Adv)/~~δ(V1)~~ | ~~δ(V1)~~ | δ(N1) | δ(Conj)/δ(V2) | δ(V1)\[δ(V2)/δ(N2)] | δ(N2) | | | |
| Step 5 | δ(V1)/~~δ(N1)~~ | | ~~δ(N1)~~ | δ(Conj)/δ(V2) | δ(V1)\[δ(V2)/δ(N2)] | δ(N2) | | | |
| Step 6 | ~~δ(V1)~~ | | | δ(Conj)/δ(V2) | ~~δ(V1)~~\[δ(V2)/δ(N2)] | δ(N2) | | | |
| Step 7 | | | | δ(Conj)/δ(V2) | δ(V2)/~~δ(N2)~~ | ~~δ(N2)~~ | | | |
| Step 8 | | | | δ(Conj)/~~δ(V2)~~ | ~~δ(V2)~~ | | | | |
| | | | | δ(Conj) | | | | | |

Fig 16. A Unified Representation

Here, the head-dependent relations can be read off from the δ functions of the categories that are in specific functor-argument relations. Likewise, the constituency relations can be understood from the way the categorial functor-argument relations are defined (that is, from the categorial cancellations), as also indicated by the arrows in Figure 16. The subsequent sections delve into the syntactic consequences and computational underpinnings of the unified representation, which are crucial for a comprehensive understanding of its nature and form.

## 5.     Syntactic Consequences of the Unified Representation

If we turn to the usefulness of the unified system of representation for various syntactic phenomena, there is a great deal of variation in the theoretical treatments of different syntactic phenomena in theoretical syntax. Varied solutions for the analysis of distinct syntactic phenomena might give an initial impression that a theoretically consistent solution for those phenomena is difficult. However, the present unified representation shows that it is in fact possible to arrive at a theoretically consistent account without deviating from the conceptual framework of the grammar formalisms. This section demonstrates how the unified representation sheds light on complex syntactic phenomena such as long-distance dependency/unbounded dependency, small clauses, complex predicates in terms of the integrated and unified representation of constituency, head-dependent and functor-argument relations. This integrated and unified representation captures information about word order, phrase/constituent structure, phrase structure rules, grammatical categories of words/morphemes, hierarchical structure in the PSG trees, etc., apart from direct head-dependent relations and functor-argument relations. Thus, the integrated and unified representation comprehensively captures both formal and functional aspects of syntactic structure.

If we focus on long-distance dependency, Head-Driven Phrase Structure Grammar (HPSG) uses feature-passing, that is, feature structures to represent the synsem (syntactic-semantic) properties of linguistic expressions. When a constituent is dislocated/moved to a higher position in the structure, its feature is passed up to the higher node, allowing the higher node to bind to the moved constituent. For example, consider the following question:

2.     *Which novel did John say that Mary read?*

Here, the wh-phrase *which novel* is moved from its base position in the embedded clause *Mary read which novel* to the higher position in the matrix clause *John said which novel Mary read* (Boeckx, 2008; Stroik, 2009; see for a different view, Putnam & Chaves, 2020). As per HPSG's analysis, the embedded clause has a feature structure that includes a feature indicating that the object of the clause is *which novel that Mary read*. When the relative pronoun is moved to the higher position in the matrix clause, its feature is passed up to the higher node, allowing the higher node to bind to the moved constituent. Overall,

HPSG's account of long-distance dependencies involves feature passing and co-indexation, which allow constituents to be bound across multiple levels of the syntactic structure. Now, if we zero in on to the unified representation's depiction of non-locality in long-distance dependencies, there is no need to specially depict the 'gap site' for the filler '*Wh*-word'. When the CG formulae are rewritten in terms of the DG functions, the formula for the verb read at the 'gap site' clearly depicts that the 'filler' is the argument corresponding to the predicate and is present to the left of the predicate/verb. Hence this obviates the need for explicitly depicting special movement operations such as (Internal) Merge, as in movement-based approaches such as Minimalist Syntax. Therefore, it also preserves the meaning of the sentence. Also, in HPSG, there is an explicit mention of SLASH, LOCAL/NON-LOCAL. Here, in this unified representation, no such extra notations are introduced.

So, the features of the syntactic phenomenon concerned, namely, non-local dependency, can be accounted for without any additional assumptions/notations and constraints in the unified representation. The properties of the word that are supposed to be in the gap are captured by the unified head-dependent and functor-argument relations. Hence, in such unbounded dependencies, though a subject or a complement may be missing, the dependent is depicted in the CG formulae, as rewritten in terms of the DG functions. So, the word missing in the gap is in fact accounted for, it is present in the formulae if not depicted. The abbreviated form of the unified representation for (2) is shown below. The assumption in (3) is that the category of the subject argument *Marie* is already cancelled by that category of *read* such that the category of the combined form indicates the meaning of *Marie read* and looks for the object argument.

3. *Which novel ... read?*
$\delta(N)$     $\delta(N)\backslash\delta(V)$

Now let us turn to the case of raising and control constructions. Raising and control are different ways in which a verb can be connected to its subject. With regards to raising, the subject of the embedded verb is also the subject of the subject-raising verb. The subject of the embedded clause appears to be raised/moved to the main clause without changing the meaning. Consider the sentence *He seems to be happy*. The verb *seems* appears in the main clause but its subject *he* is actually the subject of the embedded clause *to be happy*.

In PSG, raising is analyzed in terms of rules such as the 'raising rule', shown in a simplified form below (Chomsky & Lasnik, 1993; Hornstein, 2001, 2003).

4. Sentence (S): *He seems to be happy.*
   Pre-Derivational Representation: *[S seems [[he] to be happy] ]*
   Post-Derivational Representation: *[S [he] seems [ _ to be happy]]*

Also, there are more sophisticated movement-based accounts in the current literature on minimalist syntax (see Davies & Dubinsky, 2008). If we speak of HPSG, it is a non-movement-based view of raising/control. There is a type hierarchy, and the parts-of-speech and argument selection (transitive/intransitive) are represented in the form of a tree. In DG, in order to overcome the projectivity violations, raising is depicted as a dashed line between the head and the *Wh*-element (Osborne, 2019). In the unified representation, the CG formula(e) show information on whether the verb is transitive/intransitive, the number and direction of the argument(s), the category/parts of speech in a simple manner. So, the lexical descriptions are provided in the CG formulae in terms of delta functions. Here is an abbreviated unified representation.

5. *He         seems                          to                 be     happy.*
   $\delta(N)$   $(\delta(N)\backslash\delta(V))/\delta(\text{Infinitive})$   $\delta(\text{Infinitive})$ …
   $\delta(N)$   $(\delta(N)\backslash\delta(V))/\delta(\text{Infinitive})$   $\delta(\text{Infinitive})$
   $\delta(N)$   $\delta(N)\backslash\delta(V)$
               $\delta(V)$

As for control verbs, they take an embedded clause as their complement. Here, the subject of the main clause is also the subject of the embedded clause as in *He decided [PRO] to leave*. In GPSG, control is analyzed as the 'control rule' where the subject of the main clause controls the subject of the embedded clause. Here, the null pronoun 'PRO' is used to represent the subject of the embedded clause. Though there are movement-based accounts of control in the generative literature (Hornstein & Polinsky, 2010; see for discussion, Landau, 2013), the unified representation eliminates the need for an explicit 'PRO' or 'control rules'. It is noteworthy that the unified representation for both raising and control constructions will be the same, reflecting the fact that they have a common syntactic structure. Here is an abbreviated unified representation of our example of control.

6. *He   decided   to   leave.*
   $\delta(N)$ $(\delta(N)\backslash\delta(V1))/\delta(\text{Infinitive})$ $\delta(\text{Infinitive})$ …
   $\delta(N)$ $(\delta(N)\backslash(\delta(V1))/\delta(\text{Infinitive})$ $\delta(\text{Infinitive})$
   $\delta(N)$ $\delta(N)\backslash\delta(V1)$
   $\delta(V1)$

At this point, we are left with the question of how control constructions are then different from raising constructions. Here too, the unified representation can be of help. Since the functional-semantic features are different in control and raising constructions, the semantic specifications differentiating control from raising can be independently formulated in terms of lambda calculus and/or argument structuring mapping and so on. As far as the present unified system is concerned, the unified representation of syntax can be kept uniform, while the semantic specification can be formulated independently but in terms of the resources of the same unified representation.

7. Raising: $\delta(N)$ … $\delta(N)\backslash\delta(\text{Infinitive})$ …
   Control: ..……… $\delta(N)\backslash(\delta(V1)/\delta(\text{Infinitive}))$ $\delta(\text{Infinitive})$ …
   $\delta(N)$ …..$\delta(N)\backslash\delta(\text{Infinitive})$ …

It needs to be highlighted that the only representation available for raising would involve iterated wrapping between the infinitive and the N *he*, given the example in (6), and this applies to one of the semantic specifications for control as well. The intuitive idea behind the two representations in the case of control is that the N argument is part of the argument structure of both *decided* and *to-leave* in (6). Since control has two possible representations, this readily accounts for split control (Landau, 2008, pp. 298) in cases like (8).

8. *John proposed to Mary to help each other.*

Here, … $\delta(N)\backslash(\delta(V1)/\delta(\text{Infinitive}))$ $\delta(\text{Infinitive})$ … can take care of the relation between *John* and *proposed*, whereas $\delta(N)$ … $\delta(N)\backslash\delta(\text{Infinitive})$ … can apply to that between *Mary* and *to-help* …

    Now, let us consider the case of small clauses (SC). These are constructions where the subject and the predicate appear together without any linking verbs or copulas such as "is" etc. For example, consider the small clause *him intelligent* in *I consider him intelligent*. In PSG, the subject and the predicate are considered to be a type of non-

headed phrase (SC) generating different sets of rules for V-P constructions (Kayne, 1985). Here, the predicate is considered to be the head of the (small) clause (Stowell, 1983; Chomsky, 1986; Müller et. al., 2021). Thus, different linguistic frameworks have different analyses of small clauses and the unified representation helps neutralize these tensions by incorporating the basic principles of PSG, DG and CG as it can provide a common ground for different analyses. The following is the unified representation of *I consider him intelligent*.

9.     *I*     *consider*               *him*    *intelligent.*
     $\delta(N1)$    $((\delta(N1)\backslash\delta(V))/\delta(A))/\delta(N2)$    $\delta(N2)$   $\delta(A)$
     $\delta(N1)$    $(\delta(N1)\backslash\delta(V))/\delta(A)$                 $\delta(A)$
     $\delta(N1)$    $(\delta(N1)\backslash\delta(V))$
                      $\delta(V)$

Here, the underlying idea is that the verb *consider* first combines with the noun phrase *him* and then the combined expression can combine further with *intelligent* (see Kang, 1995, pp. 70-71). In each case of categorial cancellation above, we have a corresponding dependency relation: *consider* and *him* enter into such relations and then *consider* and *intelligent* enter into such relations. This preserves the syntax-semantics conformity as well, easily accounting for cases like *I consider this student and find that professor intelligent.*

As far as complex predicates are concerned, two or more predicates behave like a single predicate and are very common in Romance languages, French, etc. For example, consider the following French clause in (10).

10. *Marie a lu son livre.*
     Marie has read her book
     'Marie has read her book.' (Müller et. al., 2021, pp. 427)

In this example, the auxiliary *a* and the participle *lu* form a complex predicate. In HPSG's analysis of complex predicates, the argument structure of the first predicate is identified with the argument structure of the second predicate. The arguments for the first predicate will be the second predicate and its complement, whereas the argument for the second predicate will be its complement. Here, *lu* selects the argument *son livre* and the auxiliary *a* attracts this argument *son livre*. Hence, *son livre* is realized as the complement of *a*. In the case of the unified representation, this is clearly depicted by the CG formulae rewritten in terms of the DG functions for each of these words. So, as far as the unified representation's approach to complex predicates is concerned, it is a simplified description of the arguments of a given word (these include the subject and its complements for both the predicates). This overcomes the need for a separate specification of the predicate and its arguments. Here is the form of the unified representation of (10).

11.    *Marie*   *a*              *lu*        *son*     *livre.*

$\delta(N1)$    $(\delta(N1)\backslash\delta(Aux))/\delta(V)$    $\delta(V)/\delta(Det)$    $\delta(Det)/\delta(N2)$    $\delta(N2)$

$\delta(N1)$    $(\delta(N1)\backslash\delta(Aux))/\delta(V)$    $\delta(V)/\delta(Det)$    $\delta(Det)$

$\delta(N1)$    $(\delta(N1)\backslash\delta(Aux))/\delta(V)$    $\delta(V)$

$\delta(N1)$    $\delta(N1)\backslash\delta(Aux)$

$\delta(Aux)$

In a nutshell, the unified representation proceeds as per the predicate-argument structure. For instance, in HPSG, a linguistic expression is represented as a feature structure consisting of types, features and constraints. Unlike HPSG which distinguishes the linguistic objects (lexemes, phrases etc.) and their descriptions, the unified representation accounts for the syntactic properties of words as well as phrases in one go. This is possible because for the individual words, the CG formula is rewritten in terms of dependency functions, and also vice versa. Thus, argument structure, as well as phrase structure, is taken care of by the unified representation. This eliminates the need for introducing new constraints and explicitly specifying the properties of the linguistic objects. The unified representation specifies the categories of the lexemes/morphemes at the top and the categorial derivations correspond to the constituent structure(s). As such, an explicit mention of the constituent, for example, NP or VP is eliminated and these phrases (that is, NP, VP) are realized from the CG derivations. For any word, its part of speech and combinatorial properties are crucial. This is taken care of because of the mention of the category of the word at the top and its combinatorial properties are specified by the CG formulae written in terms of DG functions, and/or vice versa. The argument structure is the basic combinatorial information for a given predicate. Overall, the unified representation simplifies the expression of linguistic properties and eliminates the necessity for an intricate reference, as seen in HPSG. The next section will provide a discussion on the computational grounding of the unified representation.

## 6. Cognitive Computational Grounding of the Unified Representation

One may wonder how this unified representation concerns the computational grounding, if one assumes that a unified system of representation for PSG, DG and CG may not be required, for after all there can be translations from one formalism to another. The explanation is simple. The complexity of many viable solutions towards discontinuity in natural language thus rules out this possibility. Here, we discuss the need for a unified representation by considering a computational complexity approach in terms of the number of representations and translations possible in our neurocognitive system.

### 6.1 Computational complexity of multiple representations

When we speak of representations of a sentence analyzed by each of the three grammar formalisms in the brain, we may suppose that there are three micro-systems in the brain—one for PSG, DG and CG each. The key idea is to compute the number of possible representations of a given sentence from these grammar formalisms. The underlying motivation for considering three independent representations of linguistic structure in the brain comes from recent neuroscientific studies that raise the possibility of both constituency and dependency relations being represented in the brain (see Nefdt & Baggio, 2023). More empirical evidence based on fMRI studies comes from Lopopolo et al. (2020) who have shown that the left posterior superior temporal gyrus is sensitive to phrase structure constituency relations and the left anterior temporal pole and the left interior frontal gyrus are sensitive to dependency relations. The findings provide support for the fundamental idea of a unified system encompassing and accommodating the structural descriptions framed in terms of the three grammar formalisms. That is because the unified model predicts that the pieces of information from the representational properties of basic operations of PSG, DG, and CG will be integrated in neurocognitive processing and representation of linguistic structure. That this is indeed the case is compellingly suggested by Lopopolo et al. (2020, pp. 169): "… it is possible that different subregions of the left IFG support the analysis of different syntactic structures, in concert with either the left ATP or the left STG. In particular, the pars opercularis might work in concert with the left ATP in building sentence-level dependency representations, whereas the pars orbitalis performs operations related to the ones carried on in the left STG, having to do with hierarchical phrasal representations of the sentence." Thus, on the basis of correspondences between formalism-specific metrics and correlated blood oxygen level-dependent (BOLD) changes, they conclude that activity in areas sensitive to phrase structure constituency relations (especially in the left posterior superior temporal gyrus) may drive activity in the left interior frontal gyrus (which is sensitive to dependency relations) and also that activity in the left interior frontal gyrus may explain the activity in the left posterior perisylvian cortex.

Further suggestions along similar lines can be found elsewhere too (see Frank et al., 2012; Dehaene et al., 2015; Chang et al., 2020). Here, the assumption motivated by these findings is that these microsystems make the respective representations for each sentence the brain processes in each of the three grammar formalisms—which will be eventually shown to be untenable in this section. The number of possible representations§ for a given $n$ is $3^n$, based on the rule of combination. Here, $n$ refers to the number of sentences that can be processed by our brain in a given timeframe. The base is 3, because we are taking into account presences and/or absences of three grammar representations. Now, if $n = 2$, it means that 2 sentences can be considered in a given time frame. Let us consider these to be S1 and S2. In such a case, the total number of possible representations will be $3^2 = 9$. These nine representations are as follows: [$S1^{PSG}$, $S2^{PSG}$], [$S1^{PSG}$, $S2^{DG}$], [$S1^{PSG}$, $S2^{CG}$], [$S1^{DG}$, $S2^{DG}$], [$S1^{DG}$, $S2^{PSG}$], [$S1^{DG}$, $S2^{CG}$], [$S1^{CG}$, $S2^{PSG}$], [$S1^{CG}$, $S2^{CG}$], [$S1^{CG}$, $S2^{DG}$].

---

§ The focus here is on the combinations of formalisms and individual sentences. If n=1 (the unified representation (UR) itself), the number of representations would be ($1^n \times n$). If n=7, we shall have 7 combinations: [$S1^{UR}$], [$S2^{UR}$], [$S3^{UR}$], [$S4^{UR}$], [$S5^{UR}$], [$S6^{UR}$], [$S7^{UR}$].

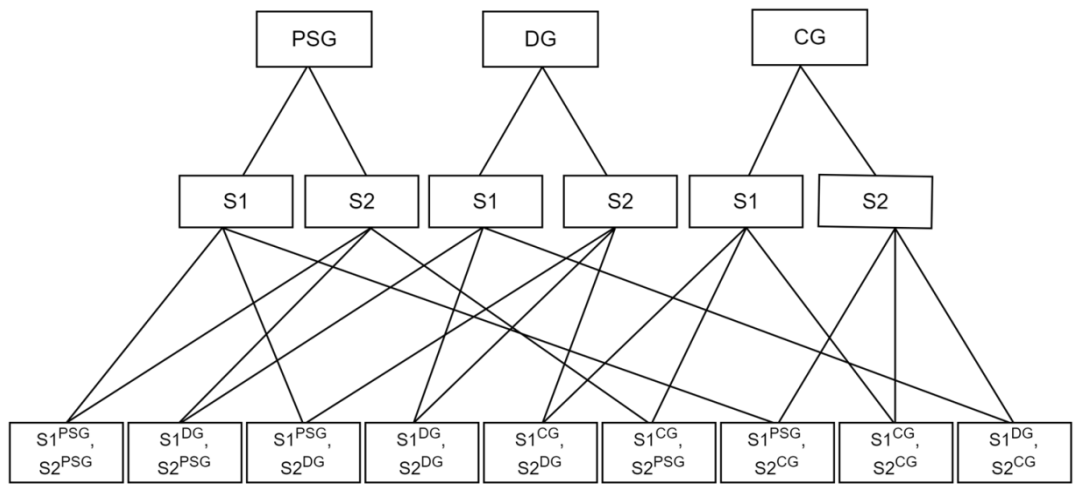

Fig 17. 9 possible representations for S1 and S2.

Miller (1956) has shown that our working memory can hold around, in fact less than, 7 items at an instant. In other words, less than 7 items or around 4 chunks can be held and processed by our brain at an instant (Cowan, 2000). This implies that our working memory for language processing has limited resources and language processing happens by recoding of materials into chunks (see Christiansen & Chater, 2016). Meanwhile, speech shadowing experiments by Marslen-Wilson (1973) have confirmed that within a window of 250-300 milliseconds, the syntactic structures of single words are available for a speaker in the neurocognitive system. In addition, the parallel processing of the syntactic and semantic structures of constituents of a sentence takes place within a window of 600-900 msec in the brain (Baggio, 2021, pp. 18). Indeed, these indicate that a phrase/constituent (about 3-4 words) is processed within a time window of 1 second and, consequently, an entire sentence will take several seconds to be processed. So, for instance, the total number of representations for the three grammar formalisms considered would be $3^7 = 2187$, if we adopt the approximate estimate that about 7 sentences can be processed by our brain in a matter of 1 minute, which is a reasonable assumption given that about 150 words can be produced per minute (Studdert-Kennedy 1986). However, it is nearly impossible for our brain to process 2187 representations per minute. Even if this number is parceled out in terms of windows of 1 second, the scenario of multiple representations does not fare any better. In view of the assumption that representations of a sequence of (whole) sentences are *not* manipulated in the working memory, we shall have about 36 representations—that is, neuronal representations of a sequence of sentence-formalism associations—for each window of 1 second. But this is clearly implausible since only a phrase/constituent, not a sequence of sentences[**], may be processed within a window of 1 second. Besides, this presents too heavy a load in processing given the incremental nature and speed of language processing over certain units of language within specific time windows (Marslen-Wilson, 1985; Rayner & Clifton, 2009; Christiansen & Chater, 2016). Thus, there must be one single unified representation for continuity and discontinuity in natural language[††]. Finally, and more crucially, there is independent psycholinguistic support for this from Bach et al. (1986) who showed that discontinuous structures triggered by cross-serial dependencies are easier to process than deeply embedded structures. This implies that both continuous structures and discontinuous structures can lead to a more or less equal processing load (see also De Vries et al., 2012).

## 6.2 Computational complexity of translations between representations of linguistic structure

We can get further insights into the computational grounding of the unified representation if we focus on translations among grammar representations. In this case, we may suppose that our neurocognitive system has to map the representation of linguistic expression (say, a phrase or a sentence) from one grammar formalism to another in order to apprehend the linguistic structures at hand. Thus, for three grammar formalisms, there has to be translations between them. The question here is: how many such translations are possible for each sentence? If we suppose that there are 3 micro-systems, each for PSG, DG and CG, let us see the number of possible mappings (M) between them. First, the number of micro-systems/grammar formalisms (N = 3) and, second, with the rule of combination used, the number of mappings/translations would be $M = N(N-1)$. Since $N=3$, $M = N(N-1) = 3(3-1) = 3 \times 2 = 6$. This means that 6 mappings/translations are possible for 3 microsystems for one single linguistic expression processed, as shown below.

---

[**] It must be recognized that the comprehension of a sentence in terms of semantic composition takes place over a window of 12-16 seconds (see Humphries et al., 2007), and hence a window of 1 second does not make the right cut for a sequence of sentences.

[††] The asymptotic growth rate will be much slower when the base is 1 (the number of formalisms is 1, that is, the UR itself). Here the assumption is that the complexity exists and will be much lower and within the bounds of human's processing capacity. Also, the possibility of having a very large *n* when the base is 1 is rare/nil, precisely because no human can perhaps process 50 or 100 sentences within a minute.

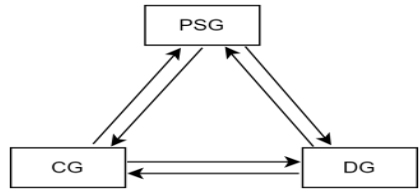

Fig 18. Possible mappings/translations between PSG, DG and CG representations

Third, if the number of sentences processed in a minute is *k*, then the total possible translations/mappings for *k* sentences is $k \times M$. If the number of sentences is 3, then the total possible mappings would be $3 \times 6 = 18$. If we consider a window of 1 minute, we shall have 7 (sentences) $\times 6 = 42$ possible mappings for 7 sentences. Also, if a window of 1 second is considered, only a phrase or constituent can be taken into account, given the estimates discussed in section 6.1. Even then, we shall have $1 \times 6 = 6$ mappings. Given that for any relationship/mapping to be established between a single constituency relation and a single dependency relation at least a phrase or a constituent needs to be processed (see Lopopolo et al., 2020), it is hard to imagine that 6 mappings can be processed for any single phrase/constituent in a matter of about 1 second. In all, the number of mappings grows too quickly with the number of phrases/constituents, and in view of the neurobiological constraints driven by the energy resources of the brain, it is highly implausible that too many mappings are established for a single sentence.

## 7 Conclusion

The unified representation makes at least three contributions to the existing literature on discontinuity. First, it shows that one can account for discontinuity in natural language without introducing any extra constraints/assumptions within the existing frameworks. Second, it has established that it is indeed possible to integrate constituency, head-dependent and functor-argument relations. Third, it demonstrates that the key ingredient towards a solution on discontinuity in natural language needs to involve the computational and neurocognitive aspects. In sum, this unified representation offers a theoretical and computational explanation of *how* discontinuous sentences are represented in our neurocognitive system.

Chomsky, N., & Lasnik, H. (1993). The theory of principles and parameters. In J. Jacobs, A. Stechow, W. Sternefeld & T. Vennemann (Eds.), *Halbband: An international handbook of contemporary research* (pp. 506-569). Berlin: De Gruyter Mouton.

Chomsky, N. (1995). Bare phrase structure. In H. Campos & P. Kempchinsky (Eds.), *Evolution and revolution in linguistic theory: Essays in honor of Carlos P. Otero* (pp. 51–109). Washington, DC: Georgetown University Press.

Christiansen, M. H., & Chater, N. (2016). *Creating language: Integrating evolution, acquisition, and processing*. Cambridge, Massachusetts: MIT Press.

Citko, B. (2011). *Symmetry in syntax: Merge, move and labels*. Cambridge: Cambridge University Press.

Cowan, N. (2000). The magical number 4 in short term memory: A reconsideration of mental storage capacity. *Behavioral and Brain Sciences*, 24, 87-185.

Davies, W. D., & Dubinsky, S. (2008). *The grammar of raising and control*. Oxford: Blackwell.

Debusmann, R. (2000). *An introduction to dependency grammar*. Hausarbeitfur das Hauptseminar Dependenzgrammatik So Se99 (Saarbrücken: Universitatdes Saarlandes).

Dehaene, S., Meyniel, F., Wacongne, C., Wang, L., & Pallier, C. (2015). The neural representation of sequences: From transition probabilities to algebraic patterns and linguistic trees. *Neuron*, 88(1), 2-19.

de Marneffe, M., & Nivre, J. (2019). Dependency grammar. *Annual Review of Linguistics*, 5, 197-218.

De Vries, M., Petersson, K. M., Geukes, S., Zwitserlood, P., & Christiansen, M. H. (2012). Processing multiple non-adjacent dependencies: Evidence from sequence learning. *Philosophical Transactions of the Royal Society B: Biological Sciences*, 367, 2065–2076.

Donohue, C., & Sag, I. A. (1999). Domains in Warlpiri. *In Sixth International Conference on HPSG–Abstracts, 04–06 August 1999* (pp. 101–106). Edinburgh: University of Edinburgh.

Dowty, D. R. (1996). Towards a minimalist theory of syntactic structure. In H. Bunt & A. Horck (Eds.), *Discontinuous constituency* (pp. 11–62). Berlin: Mouton de Gruyter.

Dras, M., Chiang, D., & Schuler, W. (2004). On Relations of Constituency and Dependency Grammars. *Research on Language and Computation*, 2, 281–305.
36